%% file: acl2020.tex
\documentclass[11pt,a4paper]{article}
\usepackage[hyperref]{acl2020}
\usepackage{times}
\usepackage{latexsym}

\usepackage{mathtools}
\usepackage{float}
\usepackage{xcolor}
\usepackage{pgfplots}
\usepackage{amsmath}
\usepackage{bbm}
\usepackage{xspace}
\usepackage{url}
\usepackage[font=small,skip=1pt]{caption}
\newcommand{\secref}[1]{\S\ref{#1}\xspace}

\newcommand{\figref}[2][]{Figure#1~\ref{#2}\xspace}
\newcommand{\equref}[2][]{Equation#1~\ref{#2}\xspace}
\newcommand{\method}[1]{\texttt{#1}\xspace}
\newcommand{\dataset}[1]{\textsc{#1}\xspace}
\newcommand{\pdtb}{\dataset{PDTB}}

\usepackage{microtype}

\aclfinalcopy 
\def\aclpaperid{17} 

\title{Learning Causal Bayesian Networks from Text}

\author{Farhad Moghimifar, \bf{Afshin Rahimi}, \bf{Mahsa Baktashmotlagh} \and \bf{Xue Li} \\
         The School of ITEE \\
         The University of Queensland, Australia\\
         \tt \{f.moghimifar,a.rahimi,m.baktashmotlagh\}@uq.edu.au\\
         \tt xueli@itee.uq.edu.au\\
         }

\date{}

\begin{document}
\maketitle

\input{0-abstract.tex}

\section{Introduction}
\label{sec:intro}
\input{1-intro.tex}

\section {Related Works}
\label{sec:related}

\input{2-relatedwork.tex}


\section {Methodology}
\label{sec:method}
\input{4-method.tex}

\section{Experimental Results}
\label{sec:experiment}
\input{5-experiments.tex}

\section{Conclusion}
\label{sec:conclusion}
\input{6-conclusion.tex}

\bibliography{ref.bib}
\bibliographystyle{acl_natbib}

\end{document}

%% file: 0-abstract.tex
\begin{abstract}
Causal relationships form the basis for reasoning and decision-making in Artificial Intelligence systems. To exploit the large volume of textual data available today, the automatic discovery of causal relationships from text has emerged as a significant challenge in recent years. Existing approaches in this realm are limited to the extraction of low-level relations among individual events. To overcome the limitations of the existing approaches, in this paper, we propose a method for automatic inference of causal relationships from human written language at conceptual level. To this end, we leverage the characteristics of hierarchy of concepts and linguistic variables created from text, and represent the extracted causal relationships in the form of a Causal Bayesian Network. Our experiments demonstrate superiority of our approach over the existing approaches in inferring complex causal reasoning from the text.
\end{abstract}

%% file: 1-intro.tex
Causation is a powerful psychological tool for human to choreograph his surrounding environment into a mental model, and use it for reasoning and decision-making. However, inability to identify causality is known to be one of the drawbacks of current Artificial Intelligence (AI) systems \citep{lake2015human}. 
Extraction of causal relations from text is necessity in many NLP tasks such as question answering and textual inference, and has attracted a considerable research in recent years \cite{wood2018challenges,zhao2018causaltriad,zhao2017constructing,ning2018joint,rojas2017causal}. 
However, the state-of-the-art methods are limited to the identification of causal relations between low-level individual events \citep{dunietz2017automatically, hidey2016identifying,mirza2016catena} and fail to capture such relationships at conceptual level. 
Furthermore, relying on linguistic features limits the identification of causal relations to those whose cause and effect are located in the same sentence or in consecutive sentences.

In this paper, we propose a method for extracting concepts and their underlying causal relations from written language. 
Furthermore, to leverage the extracted causal information, we represent the causal knowledge in the form of a \textbf{C}ausal \textbf{B}ayesian \textbf{N}etwork (CBN). 
Having this tool enables answering complex causal and counter-factual questions, such as: \textit{How psychotherapy can affect the patient's emotion?}, or \textit{What would happen if instead of medicine X, medicine Y was prescribed?}

The contribution of this paper is three-fold. Firstly, we focus on identifying causal relation between concepts (e.g. \textit{physical activity} and \textit{health}). 
Secondly, We propose a novel method to represent the extracted causal knowledge in the form of a Causal Bayesian Network, enabling easy incorporation of this invaluable knowledge into downstream NLP tasks. Thirdly, we release \dataset{PsyCaus} dataset which can be used to evaluate causal relation extraction models in the domain of psychology \footnote{https://github.com/farhadmfar/psycaus}. In addition, our proposed method identifies causality between concepts independent of their locations in text, and is able to identify bi-directional causal relations between concepts, where two concepts have causal effect on each other. By aggregating linguistic variable, we construct a hierarchy where each variable, e.g. delusional disorder, lies under its related concept, e.g. disorder. This hierarchical and inheritance structure allows for the inference of causal relations between concepts that are not directly discussed in the text.

In order to evaluate our proposed method, we gathered a corpus of psychological articles 
The experimental results shows that the proposed method performs significantly better than the state-of-the-art methods.


%% file: 2-relatedwork.tex
Identification of causality in NLP is not trivial as a result of language ambiguity. Hence, most current approaches focus on verb-verb, verb-noun, and noun-noun relations. The explicit relations are often captured with narrow syntactic and semantic constructions~\citep{do2011minimally, hendrickx2009semeval, mirza2016catena, hidey2016identifying} which limits their recall. To go beyond surface form constructions few works have proposed neural models~\citep{martinez2017neural, dasgupta2018automatic, zhao2017constructing} covering wider causal constructions. However, most works don't go beyond extracting causality between adjacent events, and so lack the ability to capture causality in non-adjacent concept level, e.g. \emph{genetics} and \emph{hallucination}. Therefore, in this paper we propose a model for identifying causality between concepts, independent of their location, and represent the causal knowledge in form of a Causal Bayesian Network.

%% file: 4-method.tex
Given the input, in form of human written language, we aim to extract the causal relation between concepts and represent the output in form of a Causal Bayesian Network. Hence, we split this task into three sub-tasks: extracting linguistic variables and values, identifying causality between extracted variables, and creating conditional probability table for each variable. In the following sub-sections each of these sub-tasks are explained.

\subsection{Linguistic Variables} \label{ling}

A \textit{linguistic variable} is a variable which values are words in natural language \citep{zadeh1983linguistic}. For example, if we consider the word Age as a linguistic variable, rather than having numeric values, its values are linguistic, such as young and old. A linguistic word is specified as $(C, T(C))$ where $C$ in the name which represents the set of words or in other word the variable's name, and $T(C)$ is the set of represented words or linguistic values. In this context, a variable and corresponding value have an asymmetric relation, in which the hypernym (superordinate) implies the hyponym (subordinate). 

In order to create a Bayesian Network (BN) from text, we first need to extract linguistic variables and values from our corpus. To this end, we leverage a probabilistic method introduced by \citet{wu2012probase} to extract all possible \textit{IsA} relations from corpus.

To enhance the accuracy of causality identification and runtime performance of our model, using Formal Concept Analysis \citep{ganter2012formal}, we represent the extracted hypernym-hyponym relations in form of a hierarchy. 
In the context of our corpus, let $V$ be the set of linguistic variables and $v$ be the set linguistic values, we call the triple of $(V,v,I)$ a formal context where $V$ and $v$ are non-empty sets and $I\subseteq V \times v$ is the incidence of the context. The pair of $(V_i,v_i)$ is called a formal concept of $(V,v,I)$ if $V_i\subseteq V, v_i\subseteq v, V_i^{'}=v_i\text{ and } v_i^{'}=V_i$, where $V_i^{'}$ and $v_i^{'}$ are sets of all attributes common to $V$ and $v$, respectively. The formal concept of a given context are naturally in form of a subconcept-superconcept relation, given for formal concepts of $ (V_i,v_i) \text{ and } (V_j,v_j) \text{ of } (V,v,I): (V_i,v_i)\leq (V_2, v_2) \iff V_i \subseteq V_j (\iff v_i\subseteq v_j)$). Consequently, we can identify that every attributes in the former formal concepts are also in the latter. Hence, this set of formula gives us the hierarchy of superconcept-subconcepts. Since every link in this hierarchy implies inheritance, attributes at the higher level are inherited by lower nodes. Therefore, if a concept $V_i$ at level $n$ of our hierarchy has a causal relation with another concept $V_j$, all the subconcepts of $V_i$ at lower level $m$ (where $m<n$), also have causal relation with $V_j$.

\subsection{Identifying Causality}
The core part of this paper is to identify the cause-effect relation between concepts, or linguistic variables. In a lower-level approach, causality is usually presented by syntactic relations, where a word or a set of words implies existence of causality. For example, `cause' in `Slower-acting drugs, like fluoxetine, may cause discontinuation symptoms" indicates a causal relation. These set linguistic features can be shown either in form of a verb or a discourse relation. The Penn Discourse Tree Bank (\pdtb) \citep{prasad2008penn} contains four coarse-grained relations, comparison, contingency, expansion and temporal, in which contingency may indicate causality. There are $28$ explicitly causal marker out of 102 in \pdtb, with the barrier of causal relation. Furthermore, we leverage the sets of verbs included in \texttt{AltLexes}, such as `force' and `caused', which show causality in a sentence. Using both discourse and verb makers of causality, we create a database of cause-effect ($\Gamma$) from given sentences. To this end, each of the input sentences are split into simpler version, using dependency parser, and once any of causality markers are identified in a sentence, the stopwords from cause and effect parts are eliminated and the remaining words are stemmed. 
Having the constructed cause-effect database ($\Gamma$), the causal relation between two concepts is defined as:
\vspace{-2ex}
\begin{equation}\label{CR}
\small
\begin{aligned}[b]
&\mathrm{CR}(V_{m},V_{n}) =& \dfrac{\sum_{i=1}^{|V_m|}\sum_{j=1}^{|V_n|} \mathbbm{1}[(v_m^i,v_{n}^j) \in \Gamma ]\vec{v}_m^i\boldsymbol{\cdot} \vec{V}_m}{\sum_{i=1}^{|V_m|} \vec{v}_m^i\boldsymbol{\cdot}\vec{V}_m} \\ 
&\hspace{12ex}-& \dfrac{\sum_{j=1}^{|V_n|}\sum_{i=1}^{|V_m|} \mathbbm{1}[(v_n^j,v_{m}^i) \in \Gamma ]\vec{v}_n^j\boldsymbol{\cdot} \vec{V}_n}{\sum_{j=1}^{|V_n|} \vec{v}_n^j\boldsymbol{\cdot}\vec{V}_n}
\end{aligned}
\end{equation}

where $V_m$ and $V_n$ are two concepts or linguistic variables in the concept space $V$, and $v_{m}^i$ is the i-th value of $V_m$; the functions $r$ and $w$ are defined as below:
\vspace{-2ex}
\begin{equation}
\small
\begin{aligned}[b]
r(a,b) = 
\begin{cases}
1 & if (a,b) \in \Gamma \\
0 & if (a,b) \notin \Gamma\\
\end{cases}
\end{aligned}
\end{equation}
and
\vspace{-2ex}
\begin{equation}
\small
\begin{aligned}
w(a,b) = 1 - S_c(a,b) = 1 - \mathit{sim}(a,b)
\end{aligned}
\end{equation}
where $\mathit{sim}$ is Cosine similarity of words $a$ and $b$. The purpose of function $w$ is to measure the relevance of the value to the corresponding variable, in order to increase the influence of more relevant values. 
The output of $CR$ can be categorised as follow:
\vspace{-2ex}
\begin{equation}\label{threshold}
\small
\mathit{CR}(A,B) \in 
\begin{cases}
(\mu, 1] & \text{A cause; B effect}\\
[- \mu, \mu] & \text{no causal relationship}\\
[-1, -\mu) & \text{B cause; A effect} \\
\end{cases}
\end{equation}
where $\mu$ is a threshold given to the model as a hyper-parameter. 
\subsection{Creating Conditional Probability Distribution} \label{cpd}
Conditional Probability Distribution (CPD) shows conditional probability of corresponding values to a variables with respect to values of parents of the variable $ P(X_i | Parents(X_i))$. In order to extend the implementation of CPD for sets of linguistic variables we use \textit{Normalized Pointwise Mutual Information (PMI)} score to calculate the probability distribution~\cite{bouma2009normalized}.
\vspace{-2ex}
\begin{align}
\small
i_n(x,y) = (ln\dfrac{p(x,y)}{p(x)p(y)})/-ln(p(x,y)) \label{pmi}
\end{align}  
The reason behind using \textit{PMI} comes from Suppes' Probabilistic theory of Causality~\cite{suppes1973probabilistic}, where he mentions that possibility of an effect to co-happen with cause is higher than happening by itself. In mathematical word it can be shown as $ P(effect|cause)>P(effect)$, which can be easily written as $ \dfrac{P(\text{cause}, \text{effect})}{P(\text{cause})P(\text{effect})} > 1$, similar to \textit{PMI} for positive values.

To create a Causal Bayesian Network from textual data, let $G$ be our graphical model, and $V=\{V_1, V_2,...,V_n\}$ be the set of extracted linguistic variables (as defined in \secref{ling}) from our corpus $\zeta$. We define $Pa_{V_i}^{G} = \{Vj: Vj \xrightarrow{causal} V_i \}$, indicating set of nodes in $\zeta$ which have causal relation with $V_i$. By expressing $P$ as:
\vspace{-2ex}
\begin{equation}
\small
P(V_{1}, V_{2}, ..., V_{n}) = \prod_{i=1}^{n}P(V_i|Pa_{V_i}^{G})
\end{equation}
we can argue that $P$ factorises over $G$. The individual factor $P(V_i|Pa_{V_i}^{G})$ is called conditional probability distribution (explained in \ref{cpd}). In a more formal way, we define Causal Bayesian Network over our corpus $\zeta$ as $\beta =(G,P)$ where $P$ is set of conditional probability distributions. In addition to the aforementioned usages of a CBN, having a Causal Bayesian Networks enables the possibility of answering questions in three different layers of Association, Intervention and Counter-factual \citep{pearl2018book}. 

%

%% file: 5-experiments.tex
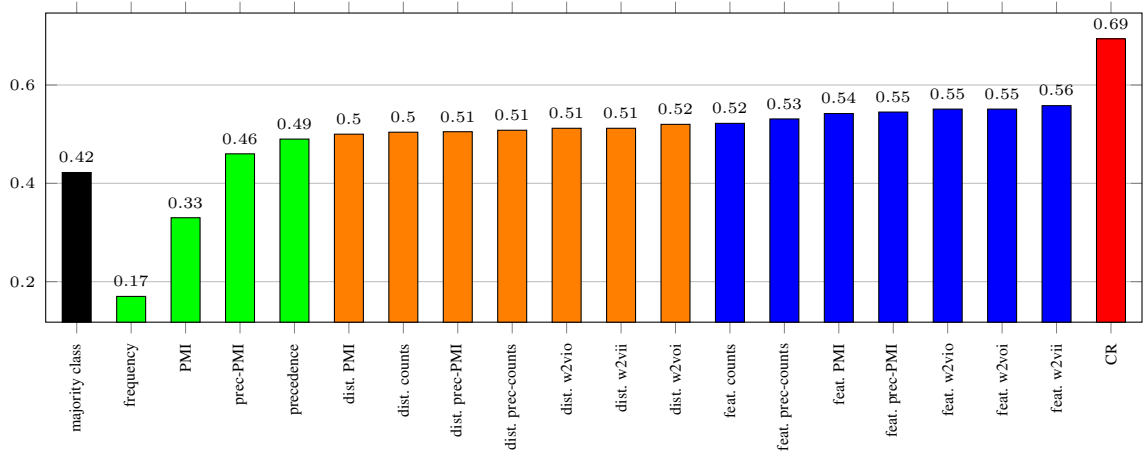
\begin{figure*}[ht!]
    \begin{tikzpicture}
      \begin{axis}[
        height=0.23\textheight,
        width=\linewidth,
        ybar,
        enlarge x limits=0.03,
        bar width =0.150in,
        bar shift=0in,
        ymajorgrids,
        x tick label style= {rotate=90,font=\tiny},
        y tick label style= {font=\tiny},
        symbolic x coords = {majority class,frequency,PMI,prec-PMI,precedence,dist. PMI,dist. counts,dist. prec-PMI,dist. prec-counts,dist. w2vio,dist. w2vii,dist. w2voi,feat. counts,feat. prec-counts,feat. PMI,feat. prec-PMI,feat. w2vio,feat. w2voi,feat. w2vii,CR},
        nodes near coords,
        every node near coord/.append style={font=\tiny},
        xtick={majority class,frequency,PMI,prec-PMI,precedence,dist. PMI,dist. counts,dist. prec-PMI,dist. prec-counts,dist. w2vio,dist. w2vii,dist. w2voi,feat. counts,feat. prec-counts,feat. PMI,feat. prec-PMI,feat. w2vio,feat. w2voi,feat. w2vii,CR},
      ]
        \addplot [ybar,fill=black] coordinates {(majority class,0.422)};
        \addplot [ybar,fill=green] coordinates { (frequency,0.17) 
                                               (PMI,0.33) 
                                               (prec-PMI,0.46) 
                                               (precedence,0.49)};
        \addplot [ybar, fill=orange] coordinates { (dist. PMI,0.500) 
                                                 (dist. counts,0.504) 
                                                 (dist. prec-PMI,0.505) 
                                                 (dist. prec-counts,0.508) 
                                                 (dist. w2vio,0.512) 
                                                 (dist. w2vii,0.512) 
                                                 (dist. w2voi,0.52) };
        \addplot[ybar, fill=blue] coordinates { (feat. counts,0.522)
                                             (feat. prec-counts,0.531)
                                             (feat. PMI,0.542)
                                             (feat. prec-PMI,0.545)
                                             (feat. w2vio,0.551)
                                             (feat. w2voi,0.551)
                                             (feat. w2vii,0.558)};
        \addplot[ybar, fill=red] coordinates {(CR,0.694)};
      \end{axis}
    \end{tikzpicture}
    \caption{The test accuracy of \emph{CR} compared with feature-based, distribution-based, heuristic methods, and the majority class.}
    \label{fig:accuracy}
\end{figure*}

In this section, we evaluate the effectiveness of our method and compare it with that of the state-of-the-art methods on a collection of Wikipedia articles related to Psychology. Each article within this collection is selected based on the terms in APA dictionary of Psychology \citep{vandenbos2007apa}. This collection contains a total number of 3,763,859 sentences. Among all possible relation between concepts, we studied 300 relationships between concepts, which were annotated by 5 graduate student from school of Psychology. Each tuple of relationship in form of $(A,B)$, were labelled as $-1,0$, or $1$, where $1$ indicates $A \xrightarrow{cause} B$, $-1$ shows that $B \xrightarrow{cause} A$, and $0$ implies no causal relations. With the overlap tuples (25\%) we measured the degree of agreement between annotators using Fleiss' Kappa measure \cite{fleiss1973equivalence}, which was around $0.67$. This indicates the reliability of our test setup. 


We compare our model to the feature-based and distribution-based methods proposed by \citet{rojas2017causal}, with different proxy projection functions, including \emph{\{ w2vii,\; w2vio,\; w2voi,\; counts,\; prec-counts,\; pmi,\; prec-pmi\}}. Furthermore, we compare our model to 
heuristic models, consisting of frequency, precedence, PMI, PMI (precedence), where in each of the models two parameters are calculated, $S_{V_i\rightarrow V_j}$ and $S_{V_i\rightarrow V_j}$, indicating $V_i\xrightarrow{cause} V_j$ if $S_{V_i\rightarrow V_j} > S_{V_i\rightarrow V_j}$ and $V_i\xleftarrow{cause} V_j$ if $S_{V_i\rightarrow V_j} < S_{V_i\rightarrow V_j}$. 

 \figref{fig:accuracy} shows the accuracy of different methods for identifying causal relationships. We observe that our method (the red bar) outperforms all other approaches with an accuracy of $0.694$. This indicates an improvement of 13\% over the state-of-the-art feature-based methods~(the blue bars), 17\% over the distribution-based approaches (the orange bars), and 20\% over the baseline methods (green bars). The baseline methods represented the worst performance, however, the accuracy achieved by precedence suggests that most of our corpus is written in form of active rather than passive voice, resulting in consequential connection between concepts.

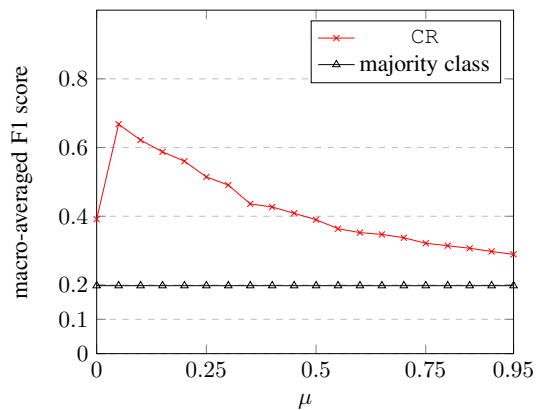
\begin{figure}[h]
\centering
\begin{tikzpicture}[scale=0.8]
\begin{axis}[
    xlabel={$\mu$},
    ylabel={macro-averaged F1 score},
    xmin=0, xmax=0.95,
    ymin=0, ymax=1,
    xtick={0,0.25,0.50,0.75,0.95},
    ytick={0,0.20,0.40,0.60,0.80,0.100},
    legend pos=north east,
    ymajorgrids=true,
    grid style=dashed,
]

\addplot[
    color=red,
    mark=x,
    ]
    coordinates {
    (0,0.391324352473933)(0.05,0.6679199455503623)(0.10,0.6218072319704211)(0.15,0.5875384178905306)(0.20,0.5602536896220934)(0.25,0.5145021707007987)(0.30,0.49079518555381557)(0.35,0.43602084410981473)(0.40,0.42694674620160145)(0.45,0.4086460065818196)(0.50,0.38994984514776543)(0.55,0.3638051135214268)(0.60,0.3523435622833213)(0.65,0.3470903608584768)(0.70,0.33743986077344745)(0.75,0.32132954704783706)(0.80,0.3142748674021741)(0.85,0.30674953865713234)(0.90,0.2974696420575269)(0.95,0.2890563119864853)
    };
    \legend{\method{CR}}

\addplot[color=black, mark=triangle] 
    coordinates {
    (0,0.19784341303328645)(0.05,0.19784341303328645)(0.10,0.19784341303328645)(0.15,0.19784341303328645)(0.20,0.19784341303328645)(0.25,0.19784341303328645)(0.30,0.19784341303328645)(0.35,0.19784341303328645)(0.40,0.19784341303328645)(0.45,0.19784341303328645)(0.50,0.19784341303328645)(0.55,0.19784341303328645)(0.60,0.19784341303328645)(0.65,0.19784341303328645)(0.70,0.19784341303328645)(0.75,0.19784341303328645)(0.80,0.19784341303328645)(0.85,0.19784341303328645)(0.90,0.19784341303328645)(0.95,0.19784341303328645)
    };
    \addlegendentry{majority class}
    
\end{axis}
\end{tikzpicture}
\caption{The macro-averaged F1 score of our proposed method on the development of \dataset{PsyCaus} with different values of $\mu$ (\equref{CR}), compared with macro-averaged F1 score of the majority class model.}
\label{figure:mu}
\end{figure}

To analyse the sensitivity of our method to the threshold $\mu$ in \equref{CR}, we trained the model on \dataset{PsyCaus}'s training set($D_{\text{tr}}$),
and analysed the development set performance in terms of macro-averaged F1 with a range of values $[0, 1)$ for $\mu$. As shown in \figref{figure:mu}, F1 score reaches the maximum of 0.66 with $\mu = 0.05$, well above a random classifier.

During the annotation process, we noticed that some concepts, e.g. \textit{eating disorder} and \textit{emotion}, may have bi-directional causal relations, depending on the context ($\text{eating disorder} \xleftrightarrow {\text{cause}} \text{emotion}$). We ran our model against these examples and found out that our approach is interestingly capable of identifying these relations as well. In \equref{CR}, approximation of absolute values of both operands in the negation to one indicates bi-directional causality. While a bi-directional causal relation cannot be presented in a CBN, as it is a \emph{directed} graphical model, a decision tree can contain these types of information. In addition, some concepts, e.g. \textit{delusional disorder} and \textit{displeasure}, that have not been connected with any type of causality connectives were also accurately identified as causal relations. This is due to the hierarchical design of variable-values in our model.


 
%% file: 6-conclusion.tex
In this paper we have presented a novel approach for identifying causal relationship between concepts. This approach enables machines to extract causality even between non-adjacent concepts. Hence, a significant improvement was delivered comparing to naive baselines. Furthermore, we represented the causal knowledge extracted from human-written language in form of Causal Bayesian Network. To the best of our knowledge this representation is novel. Having a Causal Bayesian Network can empower many downstream applications, including question-answering and reasoning. Among all applications, causal and counter-factual reasoning, which can be build on top of the outcome of this paper, may address some current hallmarks of Artificial Intelligence systems.